\documentclass[sigconf,authorversion,nonacm]{acmart}
\AtBeginDocument{%
  \providecommand\BibTeX{{%
    \normalfont B\kern-0.5em{\scshape i\kern-0.25em b}\kern-0.8em\TeX}}}

\acmConference[Woodstock '18]{Woodstock '18: ACM Symposium on Neural Gaze Detection}{June 03--05, 2018}{Woodstock, NY}
\acmBooktitle{Woodstock '18: ACM Symposium on Neural Gaze Detection,June 03--05, 2018, Woodstock, NY}

\renewcommand\footnotetextcopyrightpermission[1]{}

\begin{document}

\title[Technical Development of a Semi-Autonomous Robotic Partition]{Technical Development of a Semi-Autonomous Robotic Partition}

\author{Binh Vinh Duc Nguyen}
\email{alex.nguyen@kuleuven.be}
\orcid{0000-0001-5026-474X}
\affiliation{
  \institution{Research[x]Design, \break Department of Architecture, KU Leuven}
  \streetaddress{Kasteelpark Arenberg 1 - box 2431}
  \city{Leuven}
  \country{Belgium}
  \postcode{3001}
}

\author{Andrew Vande Moere}
\email{andrew.vandemoere@kuleuven.be}
\orcid{0000-0002-0085-4941}
\affiliation{
  \institution{Research[x]Design, \break Department of Architecture, KU Leuven}
  \streetaddress{Kasteelpark Arenberg 1 - box 2431}
  \city{Leuven}
  \country{Belgium}
  \postcode{3001}
}


\begin{abstract}
This technical description details the design and engineering process of a semi-autonomous robotic partition. This robotic partition prototype was subsequently employed in a longer-term evaluation in-the-wild study conducted by the authors in a real-world office setting.
\end{abstract}



\begin{teaserfigure}
  \includegraphics[width=\textwidth]{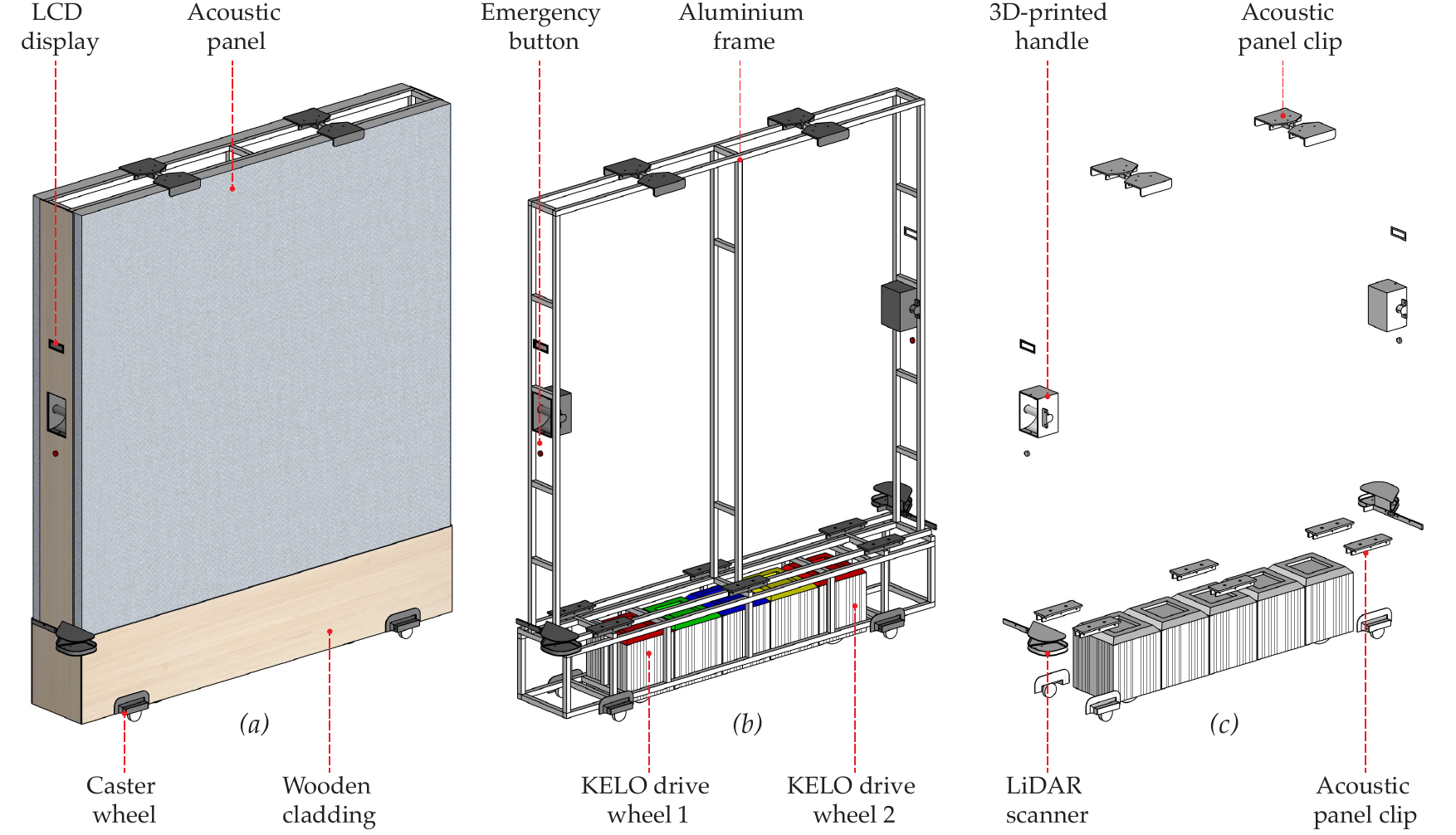}
  \caption{The robotic partition \textit{(a)}, consists of a 180x210x28cm aluminium frame that housed a customised KELO configuration of two drive wheels arranged on a single axis \textit{(b)}, with additional hardware and 3D-printed connections \textit{(c)}.}
  \label{fig:technical-wallparts}
\end{teaserfigure}

\maketitle

\section{Context}
This technical description details the design and engineering process of a semi-autonomous robotic partition. This robotic partition prototype was subsequently employed in a longer-term evaluation in-the-wild study conducted by the authors in a real-world office setting \cite{Nguyen2024}.

\section{Hardware}

The robotic partition was built on top of a customised configuration of a \(KELO\) Robile industrial robot \footnote{\(KELO\) Robile: \href{https://www.kelo-robotics.com/products/\#kelo-robile}{kelo-robotics.com}}. This customised configuration involves five \(KELO\) modules arranged on a single axis to maintain the small thickness of the robotic partition (\(28cm\)), as shown in \autoref{fig:technical-wallparts}b, including: two drive wheels, one wireless control unit, one battery with power distribution board, and one empty 'spacing' module. Because of this unusual arrangement, three main hardware solutions had to be deployed, as listed below:

\begin{itemize}
    \item \textbf{Caster wheels.} To ensure the stability of the robotic partition and prevent tipping, four casters were incorporated into the design. After evaluating various caster wheel options, I opted for a more robust choice with a load capacity of \(70kg\) each. These casters featured thick rubber wheel covers up to \(2cm\) to minimise friction when rolling on surfaces with less slip, like carpets. With a wheel diameter of \(100mm\), they provided sufficient 'span' to support the partition, enabling stable movement without overly increasing its overall footprint.

    \item \textbf{Aluminium frame.} To connect the upper sections of the partition and the caster wheels to the customised \(KELO\) robot configuration, I designed a T-slot aluminium frame, as shown in \autoref{fig:technical-wallparts}b. This choice of aluminium extrusion was based on its efficient modular assembly capability. However, upon attaching the aluminium frame to the \(KELO\) robot configuration, its movement encountered hyperstaticity \cite{Siegwart2011}, i.e. when the caster wheels lifted the drive wheels off the ground due to their rigid connection, such as while moving on an uneven surface, resulting in a disruption of robotic movement. To address this issue, I reconfigured the aluminium frame into a 'cage' that hosts the \(KELO\) robot configuration, keeping it upright yet not imposing rigidly connections, as shown in \autoref{fig:technical-hyperstatic}. This caging design allows each \(KELO\) drive wheel to flexibly shift within their own suspension system, effectively eliminating the occurrence of hyperstaticity.

\begin{figure}[h!]
  \centering
  \includegraphics[width=0.4\textwidth]{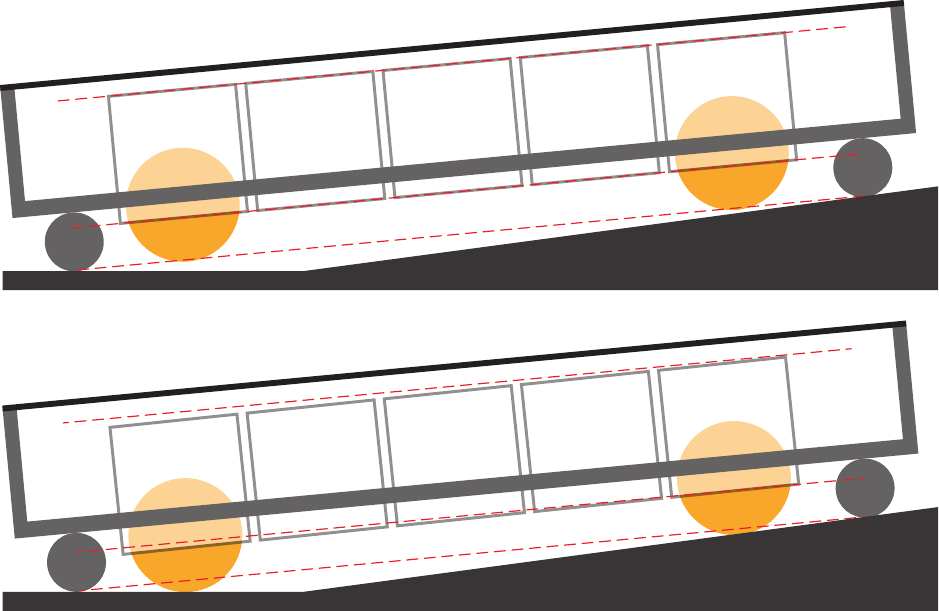}
  \caption{\textit{Top}. An exaggerated example of hyperstaticity, where drive wheels are lifted off the ground due to rigid connection with the caster wheels via aluminium frame. \textit{Bottom}. The mitigating solution, in which the aluminium frame cages the robot configuration but does not connects rigidly to it, allowing the drive wheels to flexibly shifts. }
  \label{fig:technical-hyperstatic}
\end{figure}

    \item \textbf{Drive wheel orientation.} Each \(KELO\) drive wheel was equipped with firmware that encoded its forward orientation while rotating around its \(\gamma\) angle or z-axis, referred to as its pivot encoder. Due to the single-axis configuration of the two drive wheels, it was essential to align their pivot encoders with each other, as well as with the overall movement orientation of the entire robotic partition. Consequently, it was crucial to physically connect these two wheels in a uniform orientation, consistent with the orientation of their pivot encoder to prevent potential conflicts where they might attempt to roll in different directions while both intended to move forward.
\end{itemize}

\noindent Beside the three hardware solutions above, additional hardware was integrated into the robotic partition via customised \(3D\)-printed connections to increase its affordances as shown in \autoref{fig:technical-wallparts}, including:

\begin{figure}[h!]
  \centering
  \includegraphics[width=0.5\textwidth]{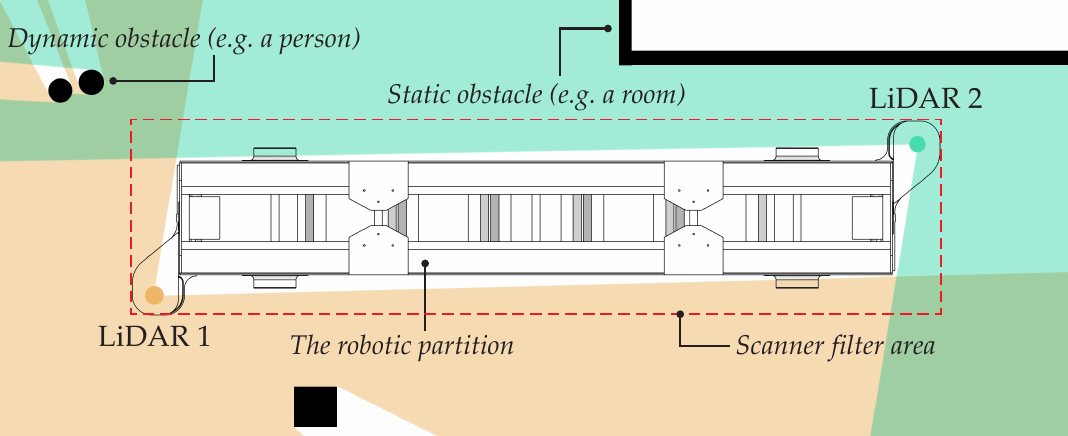}
  \caption{Top view of the robotic partition, showing the positioning of two {LiDAR} scanners, which together achieving a \(360\)-degree field of view around its perimeter.}
  \label{fig:technical-scanner}
\end{figure}

\begin{itemize}
    \item \textbf{{LiDAR} scanners.} Two \(360\)-degree {LiDAR} scanners (\(RPLIDAR\) \(A1M8{-}R6\)) were strategically mounted on the aluminium frame at extrinsic positions, accounting for the view of each {LiDAR} scanner that was covered by the robotic partition itself. By combining the data from both {LiDAR} scanners, the robotic partition achieved a \(360\)-degree field of view, extending up to a distance of \(12\) meters around its perimeter, as shown in \autoref{fig:technical-scanner}.
        
    \item \textbf{Acoustic panels.} Two acoustic panels, each measuring \(180x210cm\), were affixed to the aluminium frame to facilitate the acoustic properties of the robotic partition. These acoustic panels were not rigidly affixed to the aluminium frame but were flexibly attached using \(3D\)-printed 'clipping' elements at both the top and bottom, allowing a 'sliding' motion that is efficient for assembly.

    \item \textbf{Emergency buttons.} Two emergency buttons were situated on either side of the robotic partition, enabling an instant halt of all robotic movements upon pressing. Custom-made connections for these buttons were necessary, as they both connect to the same 'enable' pin of the battery module.

\begin{figure}[h!]
  \centering
  \includegraphics[width=0.5\textwidth]{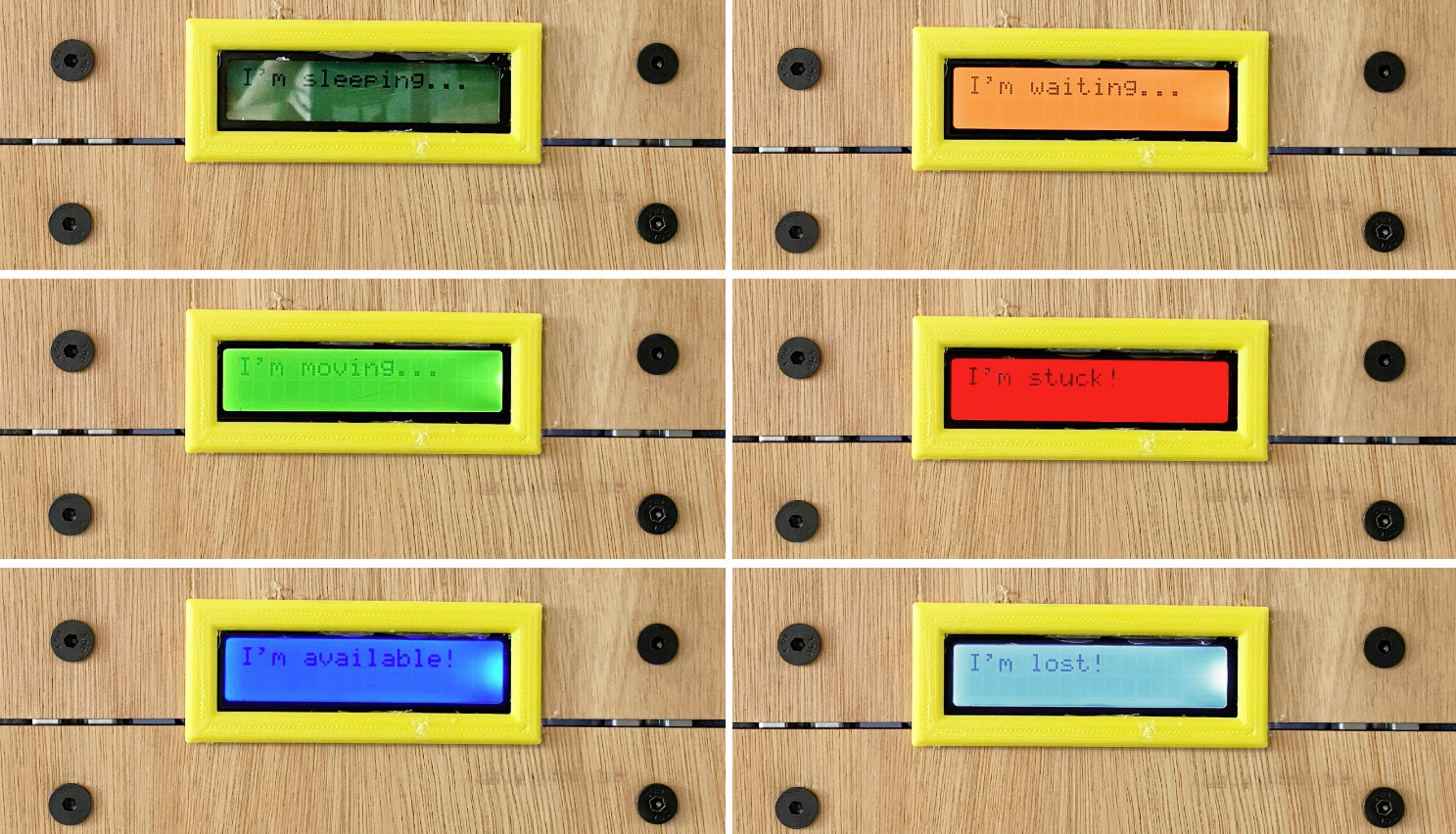}
  \caption{The six internal system states conveyed by the small integrated LCD displays on the robotic partition.}
  \label{fig:technical-display}
\end{figure}

\begin{figure*}[t!]
  \centering
  \includegraphics[width=0.9\textwidth]{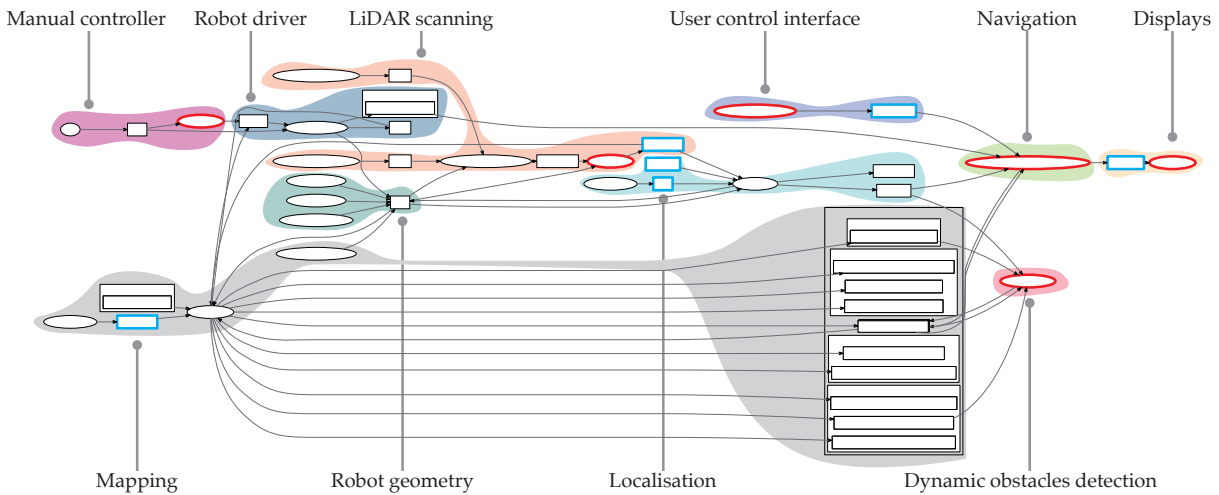}
  \caption{A diagrammatic representation of the robotic partition control software drawn by the standard \(rxgraph\). While the autonomous navigation was largely accomplished by adapting existing {ROS} libraries, six new custom modules were required.}
  \label{fig:technical-rosgraph}
\end{figure*}

    \item \textbf{Displays.} Two compact integrated \(LCD\) displays (\(Grove\) \(LCD\) \(RGB\) \(Backlight\), each connected to an \(Arduino\) \(Uno\)) were incorporated on each side of the robotic partition to convey its status to users. As illustrated in \autoref{fig:technical-display}, these navigation statuses were represented using both colours and text, encompassing descriptors such as "\textit{available}", "\textit{moving}", "\textit{waiting}" (when detecting a dynamic obstacle), or "\textit{stuck}" or "\textit{lost}" (when unable to reach its destination).

    \item \textbf{Wooden cladding and handles.} The rest of the aluminium frame was covered with wooden cladding, featuring two handles for manual manoeuvring. The wooden cladding, combined with the fabric cover of the acoustic panels, allows for the feeling of 'cosiness' normally associated to these natural materials.

\end{itemize}

\section{Software}

The semi-autonomous movement of the robotic partition was realised through a customised software built upon the {ROS} framework. As shown in \autoref{fig:technical-rosgraph}, this software integrates 13 modules adapted from existing {ROS} libraries, with six new modules that I developed exclusively for the robotic partition. Collectively, these 19 modules accomplished the main objectives below:

\begin{itemize}
    \item \textbf{Robot driver.} The robot control module utilizes the default \(kelo\_tulip\) \footnote{\(kelo\_tulip\): \href{https://github.com/kelo-robotics/kelo_tulip}{github.com}} driver library to translate velocity-based movement messages (\(geometry\_msgs/Twist\)) into the physical actuation of each drive wheel. This driver module offers flexibility in configuring the number of wheels, their physical positions relative to the centre of the robot, and their pivot encoder values as previously explained. These configuration details are stored in a dedicated file, \(config/example.yaml\). To ensure the smooth movement of this uniquely configured robot, several variables within the default velocity controller (\(kelo\_tulip/src/VelocityPlatformController.cpp\)) required optimisation. These variables encompass the minimum and maximum velocity and acceleration settings for the robot, in both linear and angular properties (six variables in total). Moreover, as each \(KELO\) drive wheel actually includes a pair of differentially-driven twin-wheels, these two individual twin wheels must undergo a brief, initial alignment phase at the outset of each robotic movement. This phase aligns them with the correct orientation around the z-axis before rolling uniformly in the commanded direction of the entire robot. Two key variables affected this alignment process: the alignment rolling speed of individual twin wheels (\(wheel\_param.pivot\_kp\)) and their alignment error threshold (\(wheel\_param.max\_pivot\_error\)). Empirical testing revealed that excessively slow \(pivot\_kp\) values in relation to the total weight of the robotic partition (i.e. less than \(0.1f\)) hindered drive wheel alignment, while excessively fast \(pivot\_kp\) values (i.e. more than \(0.6f\)) resulted in jerky motion at the start of each robot movement. Similarly, overly low \(max\_pivot\_error\) values (i.e. less than \(M\_PI{*}0.1f\)) demanded precise twin wheel alignment before the robot could commence movement, a requirement rarely met in practice and thus resulting in blocking the robot movement. Future developments of wheel drivers that can directly control their \textit{torque} values might be able to bypass these optimisation.

    \item \textbf{Manual controller.} A new module \(wall\_joy\) was developed to establish a connection between the messages from a game-pad controller to the robot driver, allowing a user to manually control the \(KELO\) robot configuration. This module is essential for debugging as well as mapping. The drive wheel orientation can be corrected by testing, for example, whether the \(Forward\) message from the game-pad controller would result in the same movement of the physical robot. If the robot moves in a wrong direction, or had trouble in aligning its orientation, the two drive wheels were probably connected with inconsistent individual orientations, or had wrong pivot encoder values.

 \begin{figure*}[h!]
  \centering
  \includegraphics[width=0.9\textwidth]{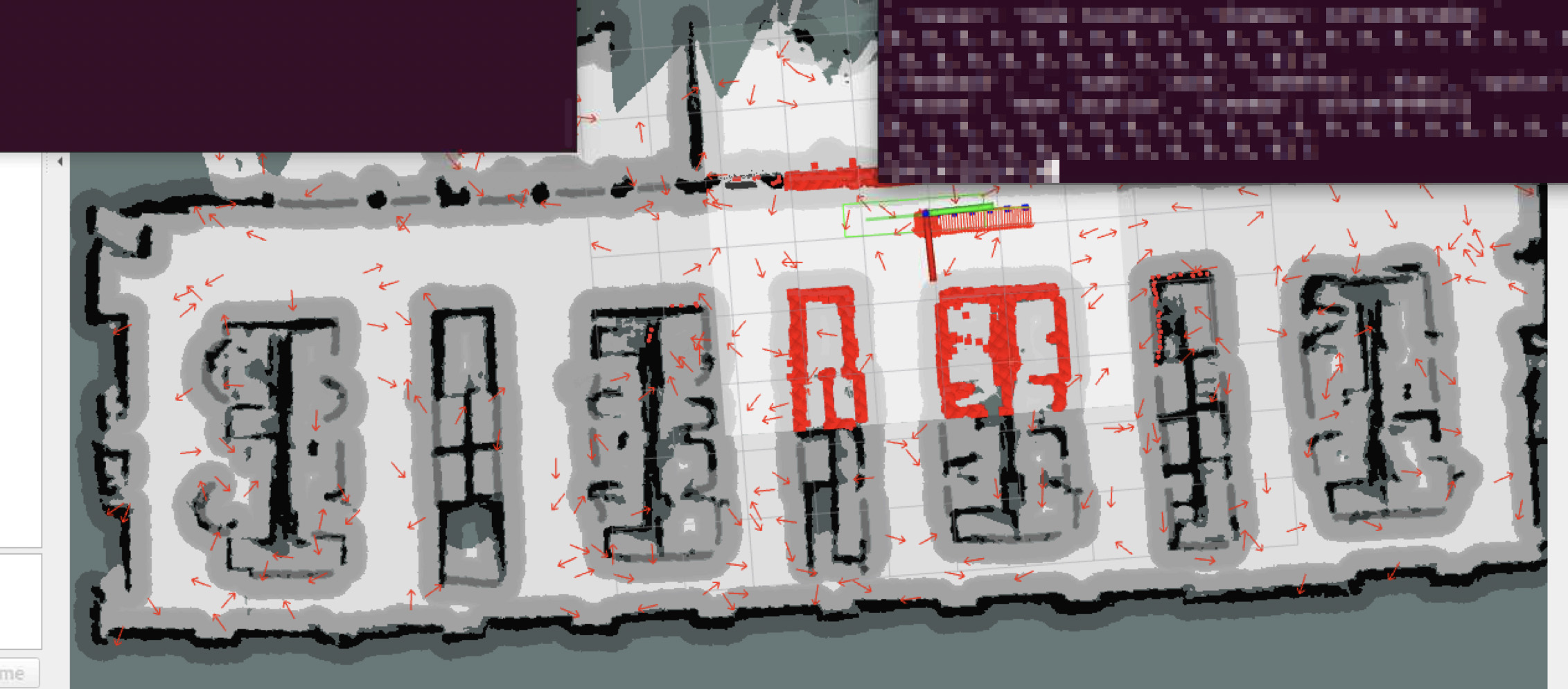}
  \caption{A screenshot of the {ROS} \(rviz\) interface with the robotic partition in action. The screenshot shows it navigating on a smooth trajectory planned by \(teb\_local\_planner\), while localising itself using \(amcl\) within a known map generated by \(Cartographer\). The {LiDAR} scanner data is shown as red squares, imposing on top of a navigation map.}
  \label{fig:technical-rosscreen}
\end{figure*}

    \item \textbf{{LiDAR} scanning.} A new module \(wall\_laser\) was developed based on two existing {ROS} libraries, \(ira\_laser\_tools\) \footnote{\(ira\_laser\_tools\): \href{http://wiki.ros.org/ira_laser_tools}{wiki.ros.org}} and \(laser\_filters\) \footnote{\(laser\_filters\): \href{http://wiki.ros.org/laser_filters}{wiki.ros.org}}, to combine the data stream from two {LiDAR} scanners and retrieve a fully \(360\)-degree \(2D\) field-of-view for the robotic partition. 
    This module also filters out the bounding box of the robotic partition itself, so that it does not register itself as an obstacle. Extensive testing revealed that the bounding box dimensions needed to be slightly larger than the actual robotic partition itself to effectively eliminate unwanted laser data. However, this larger bounding box occasionally led to the inadvertent exclusion of users who came into close proximity to the robotic partition, resulting in obstacle detection errors. Future developments may explore optimising the placement of the {LiDAR} scanners, or implementing an alternative localisation system, with the {LiDAR} scanners primarily utilised for obstacle avoidance purposes.

    \item \textbf{Mapping.} The mapping of the architectural environment was conducted using the {ROS} library of \(Cartographer\) \footnote{\(Cartographer\): \href{https://google-cartographer-ros.readthedocs.io/en/latest/}{google-cartographer-ros.readthedocs.io}}. This mapping procedure entailed moving the robotic partition with a consistent speed using a game-pad controller, allowing the {LiDAR} scanners to maintain a continuous awareness of their surroundings. Because \(Cartographer\) simultaneously triangulates odometry data from the drive wheels with point-cloud data from the {LiDAR} scanners to generate a map, any discrepancies between these data streams, such as those caused by incorrect wheel alignments, can result in sub-optimal maps that either overly compress or stretch the actual environment. As such, it is crucial to only start mapping once the physical drive wheels and their driver software are correctly set up. The resolution of the map has proven to be important, as maps with extensively high resolution (i.e. less than \(0.01\)) demand greater computational resources, leading to slower and less efficient localisation; while maps with extensively low resolution (i.e. more than \(0.1\)) tend to omit small obstacles. For this context, resolutions ranging from \(0.02\) to \(0.05\) show to be suitable. Moreover, for the robotic partition to effective avoid the obstacles that cannot be captured at the height of the physical {LiDAR} scanners, I advise employ two distinct maps. The localisation map should be generated directly by \(Cartographer\) to accurately reflect the data perceived by the {LiDAR} scanners; while the navigation map can encompass invisible obstacles superimposed onto the \(Cartographer\)-generated map using \(Photoshop\), as shown in \autoref{fig:technical-rosscreen}.

\begin{figure*}[t!]
  \centering
  \includegraphics[width=0.8\textwidth]{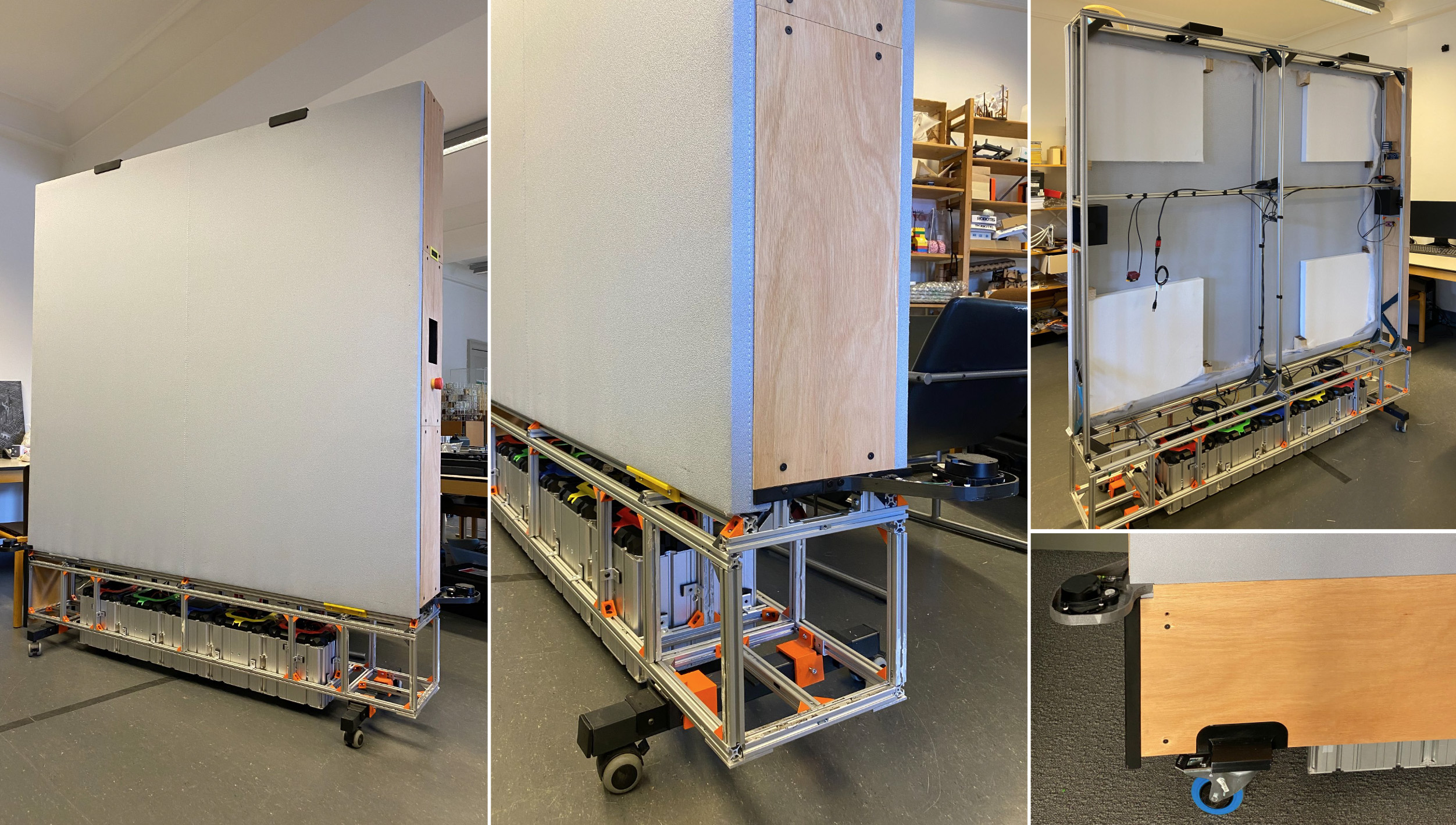}
  \caption{The gradual construction of the robotic partition, showing different options of the caster wheels and its internal acoustic materials.}
  \label{fig:technical-wall}
\end{figure*}

    \item \textbf{Localisation.} The iterative localisation of the robotic partition within the mapped environment is facilitated through the {ROS} library of \(amcl\) \footnote{\(amcl\): \href{http://wiki.ros.org/amcl}{wiki.ros.org}}. To set up this module effectively, careful configuration of its subscribed topics is required. Specifically, the \(scan\) topic should be linked to the merged and filtered {LiDAR} scanner data, while the \(map\) topic should be connected to the localisation map rather than the navigation map. Additionally, the 11 laser-model parameters should be aligned with those of the physical {LiDAR} scanners.  Initiating parameters such as \(initial\_pose\_x\), \(y\), and \(a\) (yaw) can be utilised to define the 'resting' position and orientation of the robotic partition at startup. The \(save\_pose\_rate\) parameter enables the most recent location of the robotic partition to be stored and employed as the subsequent \(initial\_pose\). Lastly, during navigation, the command service \(global\_localization\) can be invoked to restart the localisation process if the robotic partition deviates from its path and ends up in unplanned positions.

    \item \textbf{Navigation.} The autonomous navigation of the robotic partition between specified locations is achieved through the utilisation of the {ROS} library \(teb\_local\_planner\) \footnote{\(teb\_local\_planner\): \href{http://wiki.ros.org/teb_local_planner}{wiki.ros.org}}. This library, employing the trajectory optimisation method of Timed Elastic Band (\(TEB\)), is selected after extensive comparison, showing that its resulting curvature path is better suited for robots with a directional footprint, such as the robotic partition, compared to other methods like the Dynamic Window Approach (\(DWA\)) \footnote{\(dwa\_local\_planner\): \href{http://wiki.ros.org/dwa_local_planner}{wiki.ros.org}}, which is more appropriate for robots with a compact, circular footprint. To fine-tune the trajectory planner and optimise the navigation movement, extensive testing of parameter adjustments were carried out. Insights were also gathered from online forums \footnote{\(teb\_local\_planner\) \(Github\): \href{https://github.com/rst-tu-dortmund/teb_local_planner/issues}{github.com}} and previous publications \cite{Cybulski2019, Rosmann2017}. Some noteworthy considerations during this process included:
    \begin{itemize}
        \item Restricting the speed of the robotic partition within the range of \(0.2\) to \(2\) cm/s for safety, using the \(max\_vel\) parameters.
        \item Elevating the acceleration along the y-axis (\(acc\_lim\_y\)) more than the x-axis (\(acc\_lim\_x\)) to encourage sideways movement along the longer dimension of the robot, where the two drive wheels can better align with one leading and one following; rather than forward and backward movement along its shorter dimension, where the two drive wheels may encounter challenges maintaining a uniform speed and orientation.
        \item Incorporating \(viapoints\) to retrieve trajectories from the \(global\_planner\) \footnote{\(global\_planner\): \href{http://wiki.ros.org/global_planner}{wiki.ros.org}} to encourage smoother trajectory with reduced sharp angles.
        \item Aligning between \(xy\_goal\_tolerance\) and \(yaw\_goal\_tolerance\) to ensure efficient, on time navigation that still reaches its goal within acceptable error margins.
    \end{itemize}
    \noindent One limitation of the \(teb\_local\_planner\) module is its inability to determine the optimal orientation of the robotic partition within a given trajectory, such a how it should move sideways to get through a narrow corridor. To address this limitation, I developed a new module called \(wall\_move\) that publishes the optimal orientations between known locations of the robotic partition. These optimal orientations are made available to the \(teb\_local\_planner\) module through the parameter \(/move\_base/GlobalPlanner/orientation\_mode\) \footnote{\(move\_base\): \href{http://wiki.ros.org/move_base}{wiki.ros.org}}. While this solution effectively mitigated the limitation, future developments could explore the implementation of a more intelligent algorithm capable of determining suitable orientations for any trajectory automatically.

    \item \textbf{Dynamic obstacle detection.} I developed a new module called \(wall\_obstacle\) to enable the robotic partition to detect dynamic obstacles, such as passers-by, and adjust its behaviour accordingly. This module subscribes to the laser scan data obtained from \(wall\_laser\) and the current navigation trajectory planned by \(teb\_local\_planner\). It begins by checking if any dynamic obstacles are within a \(0.5m\) radius of any \(LiDAR\) scanner and then assesses whether these obstacles disrupt the planned trajectory. Depending on the available empty space around the robotic partition, the module will either initiate a navigation re-planning process to avoid the obstacle, or temporarily pause navigation for maximum \(20\) seconds or until the obstacle has cleared the area. As this algorithm highly depends on the accuracy of \(LiDAR\) scanned data, the merging and filtering of \(LiDAR\) data should be done carefully.

    \item \textbf{User control interface.} I developed two new modules to facilitate user input through a web-based control interface (\(wall\_listener\)) and to display the internal state of the robotic partition status on its two integrated LCD screens (\(wall\_screen\)). The \(wall\_listener\) module subscribes to a secure database where all user input from the web-based control interface is stored. It also publishes the navigation status of the robotic partition to the same database, providing real-time updates on the control interface. On the other hand, the \(wall\_screen\) module interprets this status information and presents it on the integrated LCD screens using a combination of text and colours.
    
\end{itemize}

\section{Conclusion}
The development process of the robotic partition, as described in this chapter, spanned approximately six months, culminating in a final prototype resembling a typical office partition, yet with the robotic capability to semi-autonomously navigate between locations within a known map. The chapter outlines various potential future technical enhancements, particularly within the mentioned modules. It is important to note that while this current prototype represents progress towards fully-autonomous spatial adaptation, important extensive further developments are necessary. These include, but are not limited to, achieving situational awareness as well as autonomously determining suitable courses of action accordingly.

\begin{acks}
The research reported in this paper is funded by grant CELSA/18/020 titled “Purposefully Controlling Mediated Architecture”.
\end{acks}

\bibliographystyle{ACM-Reference-Format}
\balance 
\bibliography{sample-base}

\end{document}